\documentclass{article}
\usepackage{ijcai18}

\usepackage{times}
\usepackage{soul}
\usepackage{url}
\usepackage[utf8]{inputenc}
\usepackage[small]{caption}
\usepackage[breaklinks=true,hidelinks]{hyperref}
\usepackage{breakcites}

\usepackage{graphicx}
\usepackage{subfigure}
\graphicspath{{figures/}{./}} 
\usepackage{psfrag}

\usepackage{color}

\title{Graph Neural Networks for Learning Robot Team Coordination}

\author{Amanda Prorok\\ 
{Department of Computer Science and Technology}\\
{University of Cambridge, UK } \\
{\texttt{asp45@cam.ac.uk}}}

\begin{document}

\maketitle

\begin{abstract}
This paper shows how Graph Neural Networks can be used for learning distributed coordination mechanisms in connected teams of robots. We capture the relational aspect of robot coordination by modeling the robot team as a graph, where each robot is a node, and edges represent communication links. During training, robots learn how to pass messages and update internal states, so that a target behavior is reached. As a proxy for more complex problems, this short paper  considers the problem where each robot must locally estimate the algebraic connectivity of the team's network topology. 
\end{abstract}

\section{Introduction}
Robot teams are becoming a de-facto solution to many of today's logistics problems (product delivery~\cite{grippa:2017}, warehousing~\cite{enright:2011}, and mobility-on-demand~\cite{pavone2012robotic}). 
Robot teams also hold the promise of delivering robust performance in unstructured or extreme environments~\cite{thayer:2001,kantor2003distributed}.
These applications hinge on algorithms that successfully and efficiently \emph{coordinate} the robots, by providing solutions to collective decision-making, formation and coverage control, and task allocation problems.

This work focuses on the problem of developing \emph{distributed} coordination mechanisms. 
To date, most distributed coordination algorithms tend to be point-solutions to very specific applications, and a lot of work goes into their design~\cite{oh:2015,garin:2010,rossi:2018}. Notably, many state-of-the art approaches rely on idealized and simplistic operational assumptions (e.g., reliability of inter-robot communications and robot homogeneity). Some of our recent work highlights the challenge of developing coordination mechanisms in \emph{heterogeneous} or \emph{faulty}~robot teams~\cite{prorok:2017TRO,saulnier:2017}: not only are these algorithms computationally hard, but also, they are difficult to design. As a consequence, we are interested in methods that more easily generate coordination mechanisms that are capable of functioning under complex operational conditions. Although some work has already been done in the domain of learning for robot team coordination~\cite{liemhetcharat:2017,amato:2016}, it is still a nascent field of research.


The goal of this paper is to apply a recent machine learning model, Graph Neural Networks (GNNs)~\cite{scarselli:2009}, to the problem of  robot team coordination. 
The GNN framework exploits the fact that many underlying relationships among data can be represented as graphs.
Although GNNs have been applied to a number of problem domains, including molecular biology~\cite{duvenaud:2015}, quantum chemistry~\cite{gilmer:2017}, and simulation engines~\cite{battaglia:2016}, they have yet to be considered within the multi-robot domain. Nevertheless, we have found that the fit is quite natural, as we capture the relational aspect of robot coordination by modeling the robot team as a graph, where each robot is a node, and edges represent communication.
This representation allows us to exploit GNNs to learn the desired coordination mechanism, where we presume that examples of the target behaviors are given to the system in a supervised learning setting.


\section{Problem and Method}
In our problem setting, robot team coordination is broken down into two main parts, \emph{(i)} inter-robot message exchange, and \emph{(ii)} robot state update. The goal is to learn both these parts. 
In a first instance (within the context of this short paper), we consider a simple problem as a proxy for more complex problems: distributed computation of the \emph{connectivity} of the robot team. 

Our notation leans on the notation in~\cite{gilmer:2017}.
We consider an undirected graph $\mathcal{G}$ with edges $\mathcal{E}$ and nodes $\mathcal{V}$. Connected nodes can pass each other messages for a duration of $T$ time-steps. Neighbors of a node $v$ are denoted by $N(v)$. Messages are denoted by $m_v^{t}$ for node $v$ at time $t$. During the message passing phase, nodes update their internal states, $h_v^{t}$. These updates are defined through a message function $M_t$ and a node update function $U_t$:
\begin{eqnarray}
m_v^{t+1} &=& \sum_{w \in N(v)} M_t (h_v^{t},h_w^{t},e_{vw}) \nonumber \\
h_v^{t+1} &=& U_t (h_v^{t},m_v^{t+1})
\end{eqnarray}
After messages have been passed, a local readout function $R_v$ returns a feature vector describing a node characteristic: 
\begin{eqnarray}
\hat{y}_v &=& R_v(h_v^T). \nonumber
\end{eqnarray}
We can also define a global readout function $\hat{y} = R (\{h_v^T | v \in G\})$ that is invariant to node permutations (graph isomorphisms).
The key point is that the functions $M_t$, $U_t$, $R_v$ and $R$ are all differentiable functions, and hence, can be learned via back-propagation. This is the premise of GNNs. 

In this work, we distribute the computation of the algebraic connectivity of the network topology of a multi-robot team (with $|\mathcal{V}|$ robots). In other words, each robot computes its own local estimate. In coordination mechanisms that rely on consensus, the algebraic connectivity is an important network property: it predicts convergence and characterizes the convergence rate. Notably, it is associated to the \emph{robustness} of network topologies~\cite{shahrivar:2015,olfati:2004}, with recent work demonstrating its effect on robot team resilience~\cite{saulnier:2017}.
The algebraic connectivity is computed by taking the second smallest eigenvalue of the graph Laplacian. Since global knowledge of the network topology is needed to compute the Laplacian, this is generally done in a centralized manner. Distributed algorithms that estimate the algebraic connectivity have previously been proposed~\cite{aragues:2012,dilorenzo:2014,poonawala:2015}. Although the details of the aforementioned estimation algorithms differ, they are all iterative approaches that lean on involved first-principles-based design. 

Our approach is to bypass the principled design of these distributed algorithms, and to estimate the algebraic connectivity ($\lambda_2$) directly via a learned coordination mechanism. Briefly stated, each node $v$ in the system estimates a local value $\lambda_{2,v}$ via the local readout function $R_v$:
\begin{eqnarray}
\hat{\lambda}_{2,v} &=& R_v(h_v^T). \nonumber
\end{eqnarray}

\section{Experiments}
\label{}
\begin{figure}[h]
\centering
\includegraphics[width=0.5\columnwidth]{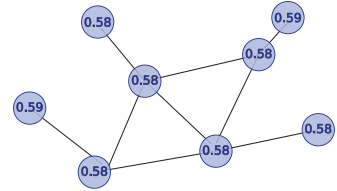}
\caption{Example of learned distributed algebraic connectivity estimation on a graph of size $|\mathcal{V}|=8$ for $T=8$. The true value is $\lambda_2=0.59$. 
Local estimates $\hat{\lambda}_{2,v}$ are superimposed on nodes.
\label{fig:results_c}}
\end{figure}

\begin{figure}[h]
\centering
\psfrag{L}[cc][][1.0][90]{Loss $\mathcal{L}_1$}
\psfrag{E}[cc][][0.9]{Epoch}
\psfrag{a}[lc][][0.65]{local, $T$=2}
\psfrag{c}[lc][][0.65]{local, $T$=4}
\psfrag{e}[lc][][0.65]{local, $T$=8}
\psfrag{i}[lc][][0.65]{global, $T$=2}
\psfrag{n}[lc][][0.65]{global, $T$=4}
\psfrag{o}[lc][][0.65]{global, $T$=8}
\includegraphics[width=0.89\columnwidth]{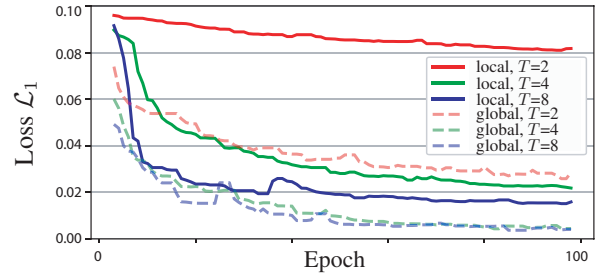}
\caption{Performance of the network, evaluated for varying durations $T$, for both a distributed model (local readout $R_v$) and a centralized model (global readout $R$).
\label{fig:results_a}}
\end{figure}

\begin{figure}[h]
\centering
\psfrag{L}[cc][][1.0][90]{Loss $\mathcal{L}_1$}
\psfrag{N}[cc][][0.9]{Number of nodes $|\mathcal{V}|$}
\psfrag{a}[lc][][0.65]{local, $T$=8}
\psfrag{e}[lc][][0.65]{global, $T$=8}
\includegraphics[width=0.88\columnwidth]{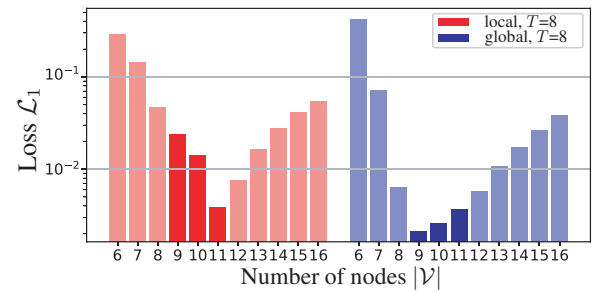}
\caption{Performance as a function of the size of the network graph. The model was trained on graph instances of size 9 to 11 nodes; the shaded bars show the performance on graph size instances not encountered during training.
\label{fig:results_b}}
\end{figure}

We adapted an implementation of GNNs available on github~\footnote{\scriptsize{\texttt{http://github.com/Microsoft/ \\ gated-graph-neural-network-samples}}}.
Our message passing function is a linear transform of the state; the state update function is handled by a GRU and the readout functions consist of a single hidden layer.
All hidden layers have size 100, and all activations are ReLUs.
The GNN is trained for a duration $T \in \{2,4,8\}$, over 100 epochs, using Adam with a learning rate of $10^{-3}$. We implement two variant GNNs: a centralized model (with global readout), and a distributed model (with local readout). 
We generated 100'000 random training examples of strongly connected graphs with $|V| \in \{9, 10, 11\}$, for which we compute the true algebraic connectivity. The default validation set comprises 10'000 graphs with $|V| \in \{9, 10, 11\}$. 
%
We train using the loss function $\mathcal{L}_2$, and our results report the error $\mathcal{L}_1$: 
\begin{eqnarray}
\mathcal{L}_2 = \frac{1}{2{|\mathcal{V}|}} \sum_i^{|\mathcal{V}|} (\hat{\lambda}_{2,i} - \lambda_2)^2,~~ \mathcal{L}_1 = \frac{1}{2{|\mathcal{V}|}} \sum_i^{|\mathcal{V}|} |\hat{\lambda}_{2,i} - \lambda_2)|. \nonumber
\end{eqnarray}
Figure~\ref{fig:results_c} shows an example of a network topology and the connectivity values predicted through our model with a local readout. 
Figure~\ref{fig:results_a} shows the average $\mathcal{L}_1$ error over validation sets, as training progresses, for three local and three global GNNs with varying messaging durations. As expected, global performs better than local, and higher $T$ perform better than lower $T$. Interestingly, increasing $T$ to 4 in the local model enables it to outperform $T=2$ in the global model.
Figure~\ref{fig:results_b} shows the ability of the models to generalize beyond the graph sizes they were trained on. As expected, the loss increases with the distance to known network sizes. For known graph size instances, the local model produces the somewhat counterintuitive result that its performance improves as graph sizes grow (the global model's behavior is the inverse).

\section{Conclusion}
This short paper demonstrates the feasibility of learning distributed coordination mechanisms for robot team coordination. 
We trained Graph Neural Networks on random network topologies, to show that accurate distributed estimation of the network connectivity is achievable. Further work will consider team coordination mechanisms beyond distributed estimation, as shown in this work.





\footnotesize
\bibliographystyle{named}
\bibliography{Bibliography}

\begin{thebibliography}{}

\bibitem[\protect\citeauthoryear{Amato \bgroup \em et al.\egroup
  }{2016}]{amato:2016}
Christopher Amato, George Konidaris, Ariel Anders, Gabriel Cruz, Jonathan~P
  How, and Leslie~P Kaelbling.
\newblock Policy search for multi-robot coordination under uncertainty.
\newblock {\em The International Journal of Robotics Research},
  35(14):1760--1778, 2016.

\bibitem[\protect\citeauthoryear{Aragues \bgroup \em et al.\egroup
  }{2012}]{aragues:2012}
Rosario Aragues, Guodong Shi, Dimos~V Dimarogonas, C~Sagues, and Karl~Henrik
  Johansson.
\newblock Distributed algebraic connectivity estimation for adaptive
  event-triggered consensus.
\newblock In {\em American Control Conference (ACC), 2012}, pages 32--37. IEEE,
  2012.

\bibitem[\protect\citeauthoryear{Battaglia \bgroup \em et al.\egroup
  }{2016}]{battaglia:2016}
Peter Battaglia, Razvan Pascanu, Matthew Lai, Danilo~Jimenez Rezende, et~al.
\newblock Interaction networks for learning about objects, relations and
  physics.
\newblock In {\em Advances in neural information processing systems}, pages
  4502--4510, 2016.

\bibitem[\protect\citeauthoryear{Di~Lorenzo and
  Barbarossa}{2014}]{dilorenzo:2014}
Paolo Di~Lorenzo and Sergio Barbarossa.
\newblock Distributed estimation and control of algebraic connectivity over
  random graphs.
\newblock {\em IEEE Transactions on Signal Processing}, 62(21):5615--5628,
  2014.

\bibitem[\protect\citeauthoryear{Duvenaud \bgroup \em et al.\egroup
  }{2015}]{duvenaud:2015}
David~K Duvenaud, Dougal Maclaurin, Jorge Iparraguirre, Rafael Bombarell,
  Timothy Hirzel, Al{\'a}n Aspuru-Guzik, and Ryan~P Adams.
\newblock Convolutional networks on graphs for learning molecular fingerprints.
\newblock In {\em Advances in neural information processing systems}, pages
  2224--2232, 2015.

\bibitem[\protect\citeauthoryear{Enright and Wurman}{2011}]{enright:2011}
John Enright and Peter~R Wurman.
\newblock Optimization and coordinated autonomy in mobile fulfillment systems.
\newblock In {\em Automated action planning for autonomous mobile robots},
  pages 33--38, 2011.

\bibitem[\protect\citeauthoryear{Garin and Schenato}{2010}]{garin:2010}
Federica Garin and Luca Schenato.
\newblock A survey on distributed estimation and control applications using
  linear consensus algorithms.
\newblock In {\em Networked Control Systems}, pages 75--107. Springer, 2010.

\bibitem[\protect\citeauthoryear{Gilmer \bgroup \em et al.\egroup
  }{2017}]{gilmer:2017}
Justin Gilmer, Samuel~S Schoenholz, Patrick~F Riley, Oriol Vinyals, and
  George~E Dahl.
\newblock Neural message passing for quantum chemistry.
\newblock {\em arXiv preprint arXiv:1704.01212}, 2017.

\bibitem[\protect\citeauthoryear{Grippa \bgroup \em et al.\egroup
  }{2017}]{grippa:2017}
Pasquale Grippa, Doris~A Behrens, Christian Bettstetter, and Friederike Wall.
\newblock Job selection in a network of autonomous uavs for delivery of goods.
\newblock {\em Robotics: Science and Systems}, 2017.

\bibitem[\protect\citeauthoryear{Kantor \bgroup \em et al.\egroup
  }{2003}]{kantor2003distributed}
George Kantor, Sanjiv Singh, Ronald Peterson, Daniela Rus, Aveek Das, Vijay
  Kumar, Guilherme Pereira, and John Spletzer.
\newblock Distributed search and rescue with robot and sensor teams.
\newblock In {\em Field and Service Robotics}, pages 529--538. Springer, 2003.

\bibitem[\protect\citeauthoryear{Liemhetcharat and
  Veloso}{2017}]{liemhetcharat:2017}
Somchaya Liemhetcharat and Manuela Veloso.
\newblock Allocating training instances to learning agents for team formation.
\newblock {\em Autonomous Agents and Multi-Agent Systems}, 31(4):905--940,
  2017.

\bibitem[\protect\citeauthoryear{Oh \bgroup \em et al.\egroup }{2015}]{oh:2015}
Kwang-Kyo Oh, Myoung-Chul Park, and Hyo-Sung Ahn.
\newblock A survey of multi-agent formation control.
\newblock {\em Automatica}, 53:424--440, 2015.

\bibitem[\protect\citeauthoryear{Olfati-Saber and Murray}{2004}]{olfati:2004}
Reza Olfati-Saber and Richard~M Murray.
\newblock Consensus problems in networks of agents with switching topology and
  time-delays.
\newblock {\em IEEE Transactions on automatic control}, 49(9):1520--1533, 2004.

\bibitem[\protect\citeauthoryear{Pavone \bgroup \em et al.\egroup
  }{2012}]{pavone2012robotic}
Marco Pavone, Stephen~L Smith, Emilio Frazzoli, and Daniela Rus.
\newblock Robotic load balancing for mobility-on-demand systems.
\newblock {\em The International Journal of Robotics Research}, 31(7):839--854,
  2012.

\bibitem[\protect\citeauthoryear{Poonawala and Spong}{2015}]{poonawala:2015}
Hasan~A Poonawala and Mark~W Spong.
\newblock Decentralized estimation of the algebraic connectivity for strongly
  connected networks.
\newblock In {\em American Control Conference (ACC), 2015}, pages 4068--4073.
  IEEE, 2015.

\bibitem[\protect\citeauthoryear{Prorok \bgroup \em et al.\egroup
  }{2017}]{prorok:2017TRO}
Amanda Prorok, M~Ani Hsieh, and Vijay Kumar.
\newblock The impact of diversity on optimal control policies for heterogeneous
  robot swarms.
\newblock {\em IEEE Transactions on Robotics}, 33(2):346--358, 2017.

\bibitem[\protect\citeauthoryear{Rossi \bgroup \em et al.\egroup
  }{2018}]{rossi:2018}
Federico Rossi, Saptarshi Bandyopadhyay, Michael Wolf, and Marco Pavone.
\newblock Review of multi-agent algorithms for collective behavior: a
  structural taxonomy.
\newblock {\em arXiv preprint arXiv:1803.05464}, 2018.

\bibitem[\protect\citeauthoryear{Saulnier \bgroup \em et al.\egroup
  }{2017}]{saulnier:2017}
Kelsey Saulnier, David Saldana, Amanda Prorok, George~J Pappas, and Vijay
  Kumar.
\newblock Resilient flocking for mobile robot teams.
\newblock {\em IEEE Robotics and Automation Letters}, 2(2):1039--1046, 2017.

\bibitem[\protect\citeauthoryear{Scarselli \bgroup \em et al.\egroup
  }{2009}]{scarselli:2009}
Franco Scarselli, Marco Gori, Ah~Chung Tsoi, Markus Hagenbuchner, and Gabriele
  Monfardini.
\newblock The graph neural network model.
\newblock {\em IEEE Transactions on Neural Networks}, 20(1):61--80, 2009.

\bibitem[\protect\citeauthoryear{Shahrivar \bgroup \em et al.\egroup
  }{2015}]{shahrivar:2015}
Ebrahim~Moradi Shahrivar, Mohammad Pirani, and Shreyas Sundaram.
\newblock Robustness and algebraic connectivity of random interdependent
  networks.
\newblock {\em IFAC-PapersOnLine}, 48(22):252--257, 2015.

\bibitem[\protect\citeauthoryear{Thayer \bgroup \em et al.\egroup
  }{2001}]{thayer:2001}
Scott~M Thayer, M~Bernardine Dias, Bart Nabbe, Bruce~Leonard Digney, Martial
  Hebert, and Anthony Stentz.
\newblock Distributed robotic mapping of extreme environments.
\newblock In {\em Mobile Robots XV and Telemanipulator and Telepresence
  Technologies VII}, volume 4195, pages 84--96. International Society for
  Optics and Photonics, 2001.

\end{thebibliography}

\end{document}